\documentclass[11pt]{article}

\usepackage[margin=1in]{geometry}

\usepackage[T1]{fontenc}
\usepackage{lmodern}
\usepackage{microtype}

\usepackage{amsmath}
\usepackage{amssymb}
\usepackage{amsfonts}
\usepackage{bm}
\usepackage{booktabs}
\usepackage{multirow}
\usepackage{array}
\usepackage{graphicx}
\usepackage{float}
\usepackage{placeins}

\usepackage{algorithm}
\usepackage{algorithmic}

\usepackage{hyperref}
\usepackage{marvosym}
\hypersetup{
  colorlinks=true,
  linkcolor=blue,
  citecolor=blue,
  urlcolor=blue,
}

\newcommand{\keywords}[1]{\par\medskip\noindent\textbf{Keywords:} #1}

\begin{document}

\begin{center}
{\LARGE\bfseries
MSSP: Multi-Scale Spatially-Constrained Partition for Unsupervised Semantic Segmentation of 3D Point Clouds\par}
\vspace{1.5em}
{\large
Zhenghao Zhang$^{1}$ \quad Xinjie Wang$^{1}$ \quad Wei Wang$^{1,*}$ \quad Jun Zhang$^{1}$ \quad Hanyun Wang$^{2}$\par}
\vspace{1.2em}
{\small
$^{1}$College of Electronic Science and Technology, National University of Defense Technology, Changsha, China\\
\vspace{0.15em}
\texttt{zhangzhenghao25@nudt.edu.cn}, \texttt{wangxinjie@nudt.edu.cn}, \texttt{wwang@nudt.edu.cn}, \texttt{zhangjun@nudt.edu.cn}\par
\vspace{0.6em}
$^{2}$School of Electronics and Communication Engineering, Sun Yat-sen University, Shenzhen, China\\
\vspace{0.15em}
\texttt{wanghanyun@mail.sysu.edu.cn}\par}
\end{center}
\vspace{0.5em}

\let\thefootnote\relax\footnotetext{\textsuperscript{*}Corresponding author.}

\begin{abstract}
3D point cloud semantic segmentation is essential for real-world spatial understanding, yet the prohibitive cost of human annotations motivates unsupervised approaches that require no labels.
Existing superpoint-based methods typically rely on spectral analysis at a fixed granularity, failing to capture the hierarchical semantic structures inherent in complex indoor scenes.
To bridge this gap, we present a Multi-Scale Spatially-Constrained Partition (MSSP) framework that combines multi-scale spectral analysis with spatially-constrained clustering.
Multi-scale spectral analysis constructs enriched superpoint descriptors across multiple clustering granularities; however, the resulting high-dimensional feature space calls for a structural prior to translate into cleaner segmentation. Spatially-constrained clustering supplies this prior by restricting superpoint merging to physically adjacent regions, imposing the spatial coherence needed for multi-scale features to be effective.
Extensive experiments on S3DIS and ScanNet show that MSSP achieves the best mIoU among unsupervised methods on the main benchmarks, with particularly significant gains on S3DIS.
Notably, our ablation reveals a regularize-then-enrich interaction: multi-scale features alone do not improve final segmentation, yet become highly effective when coupled with spatial regularization, underscoring that spatial coherence is a prerequisite for multi-scale representations in superpoint clustering.
\end{abstract}

\section{Introduction}
\label{sec:intro}

Understanding the semantics of 3D environments is essential for scene understanding, robotic navigation, and augmented reality.
Point-based networks~\cite{Qi2017CVPR,qi2017pointnetplusplus} and their subsequent extensions~\cite{wang2019dgcnn,zhao2021pointtransformer} learn directly on raw point clouds, while sparse convolutional architectures~\cite{choy2019cvpr,graham2018sparseconv} exploit efficient voxelized representations; both achieve impressive segmentation accuracy yet require large-scale point-level annotations.
A single indoor scene takes approximately 22.3 minutes to annotate~\cite{dai2017scannet}, motivating a shift toward unsupervised 3D semantic segmentation, where the goal is to discover semantically meaningful clusters without any human labeling.

To this end, GrowSP~\cite{Zhang_2023_CVPR} introduces a purely unsupervised pipeline that progressively grows superpoints to discover semantic primitives from geometric features alone, and its extension GrowSP++~\cite{zhang2026growsppp} further incorporates DINOv2~\cite{oquab2024dinov2} distillation.
PointDC~\cite{Chen_2023_ICCV} advances this direction by distilling rich 2D self-supervised features into 3D, providing stronger per-point representations for supervoxel clustering.
LogoSP~\cite{Zhang_2025_CVPR} further combines DINOv2~\cite{oquab2024dinov2} distillation with spectral analysis on a local-global superpoint graph, achieving the strongest reported results to date.
Despite this encouraging progress, all these methods pool superpoint features at a fixed clustering granularity, which cannot capture the multi-level semantic structure inherent in complex indoor scenes where objects span widely varying spatial extents.

We observe that superpoints at different clustering granularities reveal distinct semantic patterns: fine-grained clusters capture local object parts (e.g., chair legs), while coarse clusters capture whole-object semantics (e.g., entire chairs).
This suggests that concatenating features across multiple granularities should enrich the representation, analogous to how multi-scale representations~\cite{burt1983pyramid,lin2017fpn,qi2017pointnetplusplus} benefit 2D and 3D vision.
However, directly applying multi-scale features to unsupervised superpoint clustering does not improve performance on its own: multi-scale enrichment requires spatial coherence as a precondition, and without it the richer descriptors do not translate into better final segmentation.

In this paper, we propose MSSP to address this challenge.
MSSP introduces \emph{multi-scale spectral analysis} that constructs enriched superpoint descriptors at multiple clustering granularities, enabling semantic structure to be discovered at varying spatial extents.
To provide this spatial-coherence prior, we introduce \emph{spatially-constrained clustering} that restricts superpoint merging to physically adjacent regions via an adjacency graph, supplying the structural regularization needed for multi-scale features to be effective.
Our ablation study reveals a ``regularize-then-enrich'' interaction: neither component alone improves over the baseline, yet their combination yields significant gains.
The main contributions are:
\begin{itemize}
\item We advance unsupervised superpoint-based 3D semantic segmentation by introducing multi-scale spectral analysis that operates at full superpoint resolution, in contrast to the single-scale spectral analysis of LogoSP.
\item We present spatially-constrained clustering as a structural prior that supplies the spatial coherence multi-scale features require, enabling their effective use for spectral analysis.
\item We empirically demonstrate the effectiveness of MSSP on ScanNet~\cite{dai2017scannet} and S3DIS~\cite{armeni20163d}, showing that MSSP achieves the best mIoU among unsupervised methods on the two main benchmarks, with particularly significant gains on S3DIS.
\end{itemize}

\section{Related Work}
\label{sec:related}

\subsection{Supervised and Weakly-Supervised 3D Segmentation}
Point-based methods~\cite{Qi2017CVPR,qi2017pointnetplusplus,wang2019dgcnn,zhao2021pointtransformer} and sparse convolutional architectures~\cite{choy2019cvpr,graham2018sparseconv,thomas2019kpconv} achieve remarkable 3D segmentation accuracy, yet fundamentally rely on large-scale point-level annotations.
To alleviate this burden, weakly-supervised methods learn from fewer labels~\cite{liu2024u3ds3}, while 2D foundation models such as SAM~\cite{kirillov2023sam} and CLIP~\cite{radford2021clip} have been projected into 3D for open-vocabulary segmentation~\cite{jiang2024openvocab3d}.
Despite encouraging results, these approaches still require human annotations or cross-modal alignment, motivating the development of fully unsupervised alternatives.
\subsection{Unsupervised 3D Semantic Segmentation}

Unsupervised 2D methods such as IIC~\cite{ji2019iic} and PiCIE~\cite{cho2021picie} learn to cluster pixels via self-supervised features, but direct transfer to 3D remains challenging due to the domain gap~\cite{Zhang_2023_CVPR}.
GrowSP~\cite{Zhang_2023_CVPR} pioneers purely unsupervised 3D segmentation by progressively growing superpoints~\cite{landrieu2018spg} from geometric features, demonstrating that semantically meaningful primitives can emerge without any labels; its extension GrowSP++~\cite{zhang2026growsppp} further boosts discriminability by distilling DINOv2~\cite{oquab2024dinov2} features, yet still pools superpoints at a single clustering granularity.
To enrich 3D representations, PointDC~\cite{Chen_2023_ICCV} distills 2D self-supervised features into 3D; however, its supervoxel clustering is error-prone and yields inaccurate pseudo-labels.
LogoSP~\cite{Zhang_2025_CVPR} further performs spectral analysis on a local-global superpoint graph built from distilled DINOv2~\cite{oquab2024dinov2} features, where eigenvectors reveal global semantic structure, achieving the strongest results to date.
Despite this progress, all these methods pool superpoint features at a single fixed granularity, a representation that cannot resolve multi-level semantics when objects span widely varying spatial extents.

\subsection{Spectral Clustering and Multi-Scale Representation}

Spectral clustering partitions a graph by decomposing its Laplacian into eigenvectors that reveal global structure~\cite{shi2000ncut}.
In the superpoint setting, LogoSP~\cite{Zhang_2025_CVPR} applies spectral analysis to a superpoint affinity graph, where eigenvector amplitudes serve as clustering features.
Multi-scale representations are well established in 2D~\cite{burt1983pyramid,lin2017fpn} and supervised 3D~\cite{qi2017pointnetplusplus,thomas2019kpconv,hu2020randlanet} vision, yet remain unexplored in unsupervised superpoint clustering, which still pools features at this single scale.
Moreover, expanding the feature space calls for a structural prior; confining superpoint merging to physically adjacent regions via a connectivity graph provides the spatial coherence that multi-scale features require.

\section{Method}
\label{sec:method}

\begin{figure}[t]
\centering
\includegraphics[width=\textwidth]{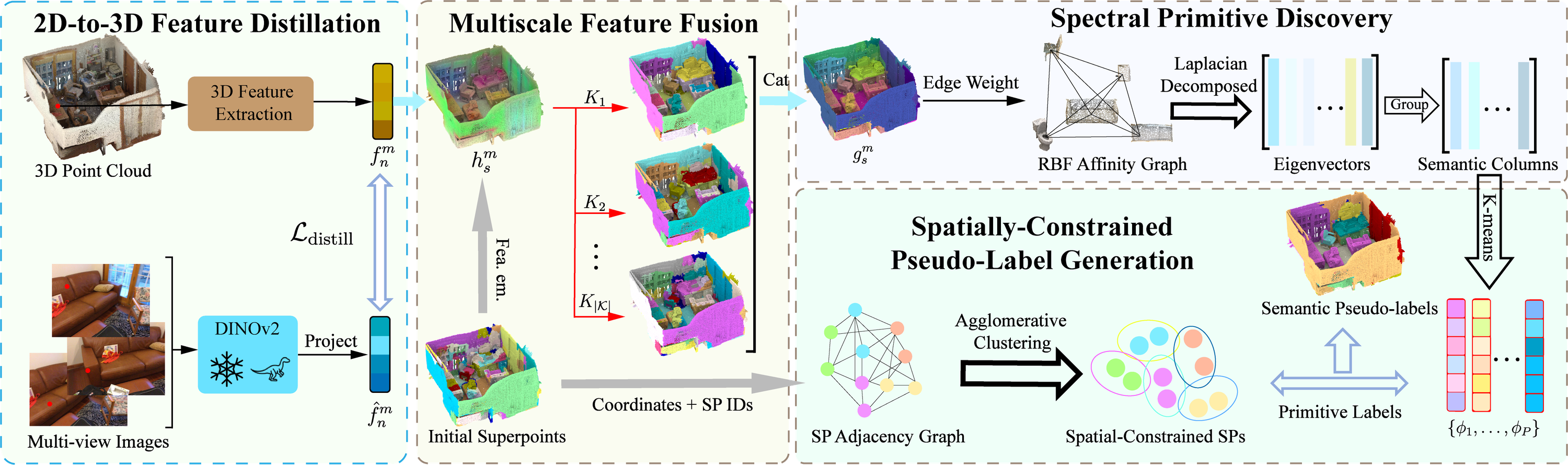}
\caption{Overview of the MSSP framework. DINOv2 features are projected to 3D and distilled into a sparse 3D backbone; the resulting per-superpoint features $h_s^m$ fork into two branches---multi-scale fusion feeds spectral primitive discovery, while spatially-constrained clustering regularizes the superpoint grouping---that converge at primitive assignment to yield pseudo-labels.}
\label{fig:pipeline}
\end{figure}

Given a dataset of $M$ scene point clouds, each scene $\mathcal{P}^m$ contains $N_m$ points with coordinates $p_n \in \mathbb{R}^3$ and color $c_n \in \mathbb{R}^3$, along with multi-view RGB-D images.
Following~\cite{Zhang_2025_CVPR,landrieu2018spg}, each scene is over-segmented into $S_m$ initial superpoints $\{\mathcal{R}_s\}_{s=1}^{S_m}$.
Our goal is to assign each point to one of $P$ semantic primitives without labels, mapped to named classes via Hungarian matching.
Figure~\ref{fig:pipeline} illustrates the overall architecture.

\textbf{2D-to-3D Feature Distillation.}
\label{sec:distill}
Following~\cite{Zhang_2025_CVPR,Chen_2023_ICCV}, a frozen DINOv2~\cite{oquab2024dinov2} ViT-S/14 backbone extracts per-pixel features from multi-view images, which are projected onto 3D points using depth and camera parameters~\cite{Zhang_2025_CVPR}.
A 3D sparse convolutional backbone (Res16FPN18~\cite{choy2019cvpr}) is trained to match the projected features $\hat{f}_n^m \in \mathbb{R}^D$ ($D{=}384$) via cosine similarity:
\begin{equation}
\mathcal{L}_{\text{distill}} = \frac{1}{N_m} \sum_{n=1}^{N_m} \left(1 - \cos(f_n^m, \hat{f}_n^m)\right)
\label{eq:distill_loss}
\end{equation}

\textbf{Superpoint Feature Extraction.}
\label{sec:sp_feat}
The distilled backbone extracts per-voxel features $f_n^m$, which are average-pooled within each initial superpoint $\mathcal{R}_s$:
\begin{equation}
h_s^m = \frac{1}{|\mathcal{R}_s|} \sum_{n \in \mathcal{R}_s} f_n^m \in \mathbb{R}^D
\label{eq:sp_feat}
\end{equation}

\textbf{Spatially-Constrained Clustering.}
\label{sec:spatial}
LogoSP~\cite{Zhang_2025_CVPR} merges superpoints with unconstrained KMeans on single-scale features, which can group spatially non-adjacent superpoints into the same cluster.
In contrast, we replace KMeans with spatially-constrained agglomerative clustering that confines each merge to physically adjacent superpoints, keeping the resulting groups geometrically coherent.

For scene $\mathcal{P}^m$, we construct a superpoint adjacency graph $\mathcal{G}^m = (\mathcal{V}, \mathcal{E})$ where $\mathcal{V} = \{1, \ldots, S_m\}$ indexes the initial superpoints.
An edge $(i, j) \in \mathcal{E}$ exists if any voxel in $\mathcal{R}_i$ is a 6-neighbor of any voxel in $\mathcal{R}_j$ in the voxelized coordinate grid:
\begin{equation}
\mathcal{E} = \left\{(i,j) \;\middle|\; \exists\, v_i \in \mathcal{R}_i,\, v_j \in \mathcal{R}_j : \|v_i - v_j\|_1 = 1 \right\}
\label{eq:adjacency}
\end{equation}
This connectivity matrix is used as the \texttt{connectivity} constraint in AgglomerativeClustering~\cite{scikit-learn}, which merges the $S_m$ initial superpoints bottom-up into approximately $C$ spatially-connected groups per scene ($C{=}80$ on S3DIS, $C{=}40$ on ScanNet).
After merging, per-voxel features are re-pooled within each merged group.
Unlike LogoSP, where these merged features directly feed spectral analysis, our multi-scale fusion (Section~\ref{sec:multiscale}) operates instead on the original initial-superpoint features, enriching them with multi-granularity context for spectral analysis.
The merged groups thus do not modify the spectral affinity graph; they serve only as the spatially-connected units on which primitive assignment later operates (Section~\ref{sec:training}).

\textbf{Multi-Scale Feature Fusion.}
\label{sec:multiscale}
Existing superpoint methods~\cite{Zhang_2025_CVPR,Zhang_2023_CVPR,Chen_2023_ICCV} pool features at a single fixed granularity, treating all superpoints identically regardless of whether they represent a small object part or a large semantic region.
In contrast, we propose to enrich superpoint descriptors by capturing semantic patterns at multiple clustering granularities using \emph{semantic granularity} as the scale parameter: fewer KMeans clusters produce broader semantic groups, while more clusters preserve finer distinctions.

Let $\{h_s^m\}_{s=1}^{S_m} \subset \mathbb{R}^D$ denote the superpoint features for scene $m$.
We construct multi-scale descriptors at $|\mathcal{K}|$ granularities $\mathcal{K} = \{K_1, \ldots, K_{|\mathcal{K}|}\}$ (e.g., $\mathcal{K} = \{80, 40, 20\}$).
For each $K \in \mathcal{K}$, we run KMeans on L2-normalized features $\hat{h}_s^m = h_s^m / \|h_s^m\|_2$ with $K$ clusters (Group), compute cluster centroids in the original feature space (Aggregate):
\begin{equation}
\mu_k^{(K)} = \frac{1}{|\mathcal{C}_k^{(K)}|} \sum_{s \in \mathcal{C}_k^{(K)}} h_s^m, \quad \mathcal{C}_k^{(K)} = \{s : \pi_K(s) = k\}
\label{eq:centroid}
\end{equation}
and assign each superpoint its normalized centroid as a contextual descriptor (Assign):
\begin{equation}
g_s^{m,(K)} = \frac{\mu_{\pi_K(s)}^{(K)}}{\|\mu_{\pi_K(s)}^{(K)}\|_2}
\label{eq:map_back}
\end{equation}
Each scale replaces the raw superpoint feature with its cluster centroid, so the descriptor reflects cluster-level context rather than the raw per-superpoint signal.
The final multi-scale feature concatenates all scales:
\begin{equation}
g_s^m = [g_s^{m,(K_1)}; g_s^{m,(K_2)}; \ldots; g_s^{m,(K_{|\mathcal{K}|})}] \in \mathbb{R}^{D \cdot |\mathcal{K}|}
\label{eq:multiscale_concat}
\end{equation}
A key difference from LogoSP is that spectral analysis now operates on all $S_m$ initial superpoints (${\sim}700$ per scene for S3DIS) rather than on the ${\sim}P$ merged superpoints, providing substantially finer spatial resolution for discovering semantic primitives.
The final primitive assignment reverts to the original single-scale features $h_s^m$, since the classifier $W$ must operate in the same feature space as the backbone output, so that the backbone can be trained end-to-end to predict these primitives.
Multi-scale features and spatially-constrained clustering are constructed independently of the same initial superpoint features $\{h_s^m\}$, yet exhibit a ``regularize-then-enrich'' interaction: the expanded multi-scale feature space calls for a spatial-coherence prior, and spatially-constrained clustering supplies this prior so that the richer descriptors converge to coherent groupings, as validated in Section~\ref{sec:ablation}.

\textbf{Spectral Primitive Discovery and Pseudo-Label Generation.}
\label{sec:training}
Following LogoSP~\cite{Zhang_2025_CVPR}, the multi-scale features $\{g_s^m\}$ undergo spectral primitive discovery.
LogoSP constructs the affinity matrix on $P$ merged superpoints with single-scale 384-dim features; in contrast, we build it on all $S_m$ initial superpoints with multi-scale $D{\cdot}|\mathcal{K}|$-dim features, providing both richer semantic structure and finer spatial resolution.
Specifically, an RBF affinity matrix is built and its normalized Laplacian decomposed into eigenvectors; each superpoint is then represented by its coefficients on the leading eigenvectors (selected by an energy threshold), forming a spectral embedding that a final KMeans partitions into $P$ semantic primitives $\{\phi_1, \dots, \phi_P\}$.
The two branches converge at \emph{primitive assignment}: each merged superpoint is matched to its nearest primitive centroid---computed in the original $D$-dimensional space---via cosine similarity, and the label propagated to its constituent points as pseudo-labels.
A linear classifier $W \in \mathbb{R}^{D \times P}$ is initialized with the L2-normalized primitive centers and frozen during training.
The backbone is trained with cross-entropy loss against the pseudo-labels, and the pseudo-label pipeline may be re-executed at a fixed interval using updated backbone features; the exact schedule is dataset-specific and detailed in Sec.~\ref{sec:implementation}.
At inference, primitive-to-class mapping uses Hungarian matching~\cite{Zhang_2025_CVPR,Chen_2023_ICCV}. Following the standard unsupervised protocol, a single global assignment is computed once per evaluation split from the predicted-vs-ground-truth co-occurrence histogram on that split, and is used only for the post-hoc naming of discovered clusters; the backbone itself is trained without any labels. For the ScanNet online benchmark, both the cluster-to-class assignment and checkpoint selection use only the labeled train+val set, and the resulting mapping is then transferred unchanged to the unlabeled online test set.

\section{Experiments and Analysis}
\label{sec:experiments}

\begin{figure}[t]
\centering
\includegraphics[width=\textwidth]{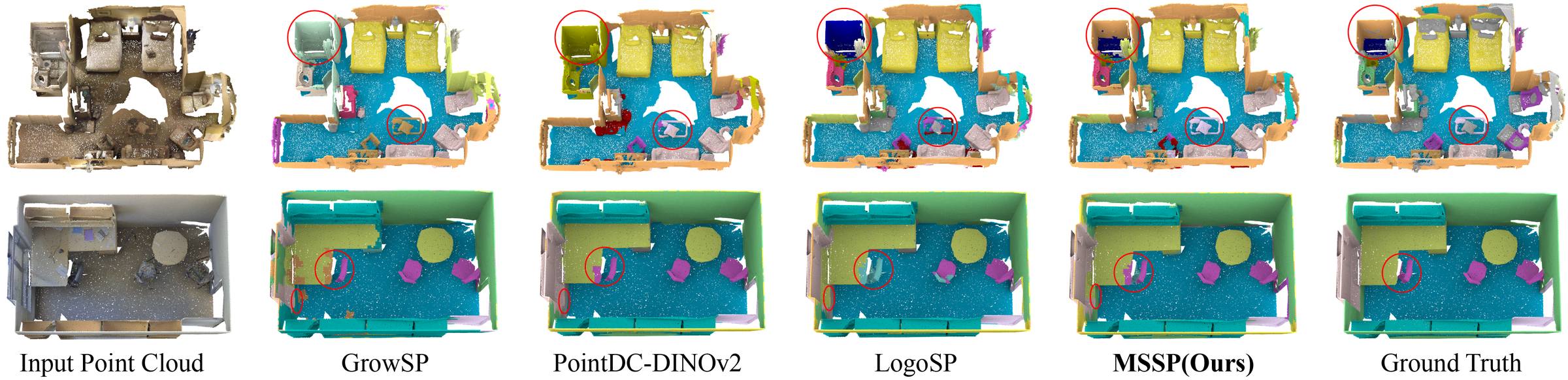}
\caption{Qualitative comparison on ScanNet (top row) and S3DIS (bottom row). From left to right: representative baseline methods, MSSP (Ours), and Ground Truth. MSSP produces more coherent and accurate segments, particularly on structurally complex classes and at object boundaries.}
\label{fig:qualitative}
\end{figure}

\subsection{Datasets and Benchmarks}
\label{sec:benchmarks}

\noindent\textbf{Datasets.}
We evaluate on \textbf{ScanNet}~\cite{dai2017scannet} and \textbf{S3DIS}~\cite{armeni20163d}.
ScanNet consists of 1,201/312/100 scenes for training, validation, and online test respectively, with 20 semantic classes.
S3DIS comprises 6 large areas with 271 rooms and 13 semantic classes. Following the standard Area~5 protocol, we train on Areas 1, 2, 3, 4 and 6, and evaluate on the held-out Area~5; we additionally report 6-fold cross-validation (leave-one-area-out, averaged over all six areas).

\noindent\textbf{Metrics.}
We report mIoU, Overall Accuracy (OA), and mean class Accuracy (mAcc).

\noindent\textbf{Baselines.}
Unsupervised: K-means, IIC~\cite{jaritz2020xmuda}, PiCIE~\cite{cho2021picie}, GrowSP~\cite{Zhang_2023_CVPR}, PointDC~\cite{Chen_2023_ICCV}, PointDC-DINOv2, LogoSP~\cite{Zhang_2025_CVPR}.
Supervised: PointNet~\cite{Qi2017CVPR}, PointNet++~\cite{qi2017pointnetplusplus}, SparseConv~\cite{graham2018sparseconv}.
GrowSP++~\cite{zhang2026growsppp} extends GrowSP by additionally distilling DINOv2 features, so it shares our 2D-pretrained setting and serves as a directly comparable baseline. GrowSP~\cite{Zhang_2023_CVPR} is the only purely-geometric baseline (no 2D pretrained features) and is marked with $^\dagger$ in Tables~\ref{tab:scannet} and~\ref{tab:s3dis}.

\subsection{Implementation Details}
\label{sec:implementation}

We use DINOv2~\cite{oquab2024dinov2} ViT-S/14 as the 2D backbone and Res16FPN18~\cite{choy2019cvpr} as the 3D backbone, outputting $D{=}384$ dimensional features.
For multi-scale analysis, $\mathcal{K} = \{80, 40, 20\}$ produces a $1152$-dim concatenated descriptor.
Following~\cite{Zhang_2025_CVPR}, eigenvector selection uses energy thresholding $\eta{=}0.99$; the number of primitives is set equal to the number of semantic classes, $P{=}20$ on ScanNet and $P{=}12$ on S3DIS.
The adjacency graph uses 6-neighbor voxel connectivity as the constraint for AgglomerativeClustering.
Unless stated otherwise, both stages use an Adam-family optimizer (Adam for ScanNet, AdamW for S3DIS) with a learning rate of $10^{-4}$, weight decay $10^{-4}$, and a PolyLR schedule; we train for 200 epochs on ScanNet and 100 on S3DIS with a batch size of 10 scenes at voxel size $0.05$\,m. KMeans uses $k$-means++ initialization with a fixed seed (\texttt{random\_state}{=}0). Pseudo-labels are refreshed every 10 epochs on S3DIS and retained from the initial assignment on ScanNet; all random seeds default to 2022, and checkpoint selection follows LogoSP~\cite{Zhang_2025_CVPR}.
The pipeline runs in two stages. In the 2D-to-3D distillation stage, the backbone is trained for 300 epochs on ScanNet and 700 on S3DIS to match the projected DINOv2 features via cosine similarity, and the distilled weights initialize the segmentation stage. During segmentation training, each scene is augmented on the fly with random rotation, translation, scaling, and elastic deformation.
All experiments run on a single NVIDIA RTX 4090.

\subsection{Quantitative Results}
\label{sec:results}

\noindent\textbf{ScanNet.}
Table~\ref{tab:scannet} reports results on ScanNet.
MSSP achieves the best OA ($70.7$, $+6.0$ over LogoSP), mIoU ($36.2$, $+0.4$), and mAcc ($51.3$, $+0.5$) among unsupervised methods.

\begin{table}[H]
\centering
\caption{Results on ScanNet dataset. Best unsupervised results in \textbf{bold}. Supervised methods report only online mIoU; ``---'' = metric not reported. $^\dagger$GrowSP does not use pretrained 2D visual features; MSSP, GrowSP++, and the other unsupervised baselines use pretrained DINO/DINOv2 features, so gains over GrowSP partly reflect stronger 2D pretraining.}
\label{tab:scannet}
\small
\setlength{\tabcolsep}{3.5pt}
\begin{tabular}{l ccc c}
\toprule
& \multicolumn{3}{c}{Val Split} & Online \\
\cmidrule(lr){2-4} \cmidrule(lr){5-5}
Method & OA(\%) & mAcc(\%) & mIoU(\%) & mIoU(\%) \\
\midrule
\multicolumn{5}{l}{\textit{Supervised}} \\
PointNet++~\cite{qi2017pointnetplusplus}  & \multicolumn{3}{c}{---} & 33.9 \\
PointCNN~\cite{li2018pointcnn}          & \multicolumn{3}{c}{---} & 45.8 \\
SparseConv~\cite{graham2018sparseconv}    & \multicolumn{3}{c}{---} & 72.5 \\
\midrule
\multicolumn{5}{l}{\textit{Unsupervised}} \\
K-means          & 10.1 & 10.0 & 3.4  & --- \\
IIC~\cite{jaritz2020xmuda}              & 27.7 & 6.1  & 2.9  & --- \\
PiCIE~\cite{cho2021picie}             & 20.4 & 16.5 & 7.6  & --- \\
GrowSP$^\dagger$~\cite{Zhang_2023_CVPR}           & 57.3 & 44.2 & 25.4 & 26.9 \\
GrowSP++~\cite{zhang2026growsppp}              & 70.5 & 48.8 & 33.2 & 32.3 \\
PointDC~\cite{Chen_2023_ICCV}           & 63.7 & ---  & 25.7 & 22.9 \\
PointDC-DINOv2~\cite{Chen_2023_ICCV}    & 64.7 & 45.0 & 29.6 & --- \\
LogoSP~\cite{Zhang_2025_CVPR}           & 64.7 & 50.8 & 35.8 & 32.7 \\
\textbf{MSSP (Ours)} & \textbf{70.7} & \textbf{51.3} & \textbf{36.2} & \textbf{34.7} \\
\bottomrule
\end{tabular}
\end{table}

The OA improvement ($+6.0$) is markedly larger than the mIoU improvement ($+0.4$): OA reflects overall point accuracy and is driven by the high-frequency structural classes that dominate ScanNet scenes, whereas mIoU averages per-class IoU and is held back by minority classes with little headroom in compact single-room layouts.
Consistent with this, MSSP also leads on the hidden online test set ($34.7$ mIoU, $+2.0$ over LogoSP), where the larger gap suggests that the spatial-coherence prior generalizes to unseen scenes.
Compared to the DINOv2-pretrained GrowSP++ ($33.2$ mIoU) and PointDC-DINOv2 ($29.6$ mIoU), MSSP improves by $+3.0$ and $+6.6$ mIoU respectively, confirming that the gain stems from our multi-scale spectral analysis rather than the 2D features alone.

\noindent\textbf{S3DIS.}
Table~\ref{tab:s3dis} reports results on S3DIS.
Under the Area~5 setting, MSSP achieves the best mIoU ($49.2$, $+2.6$ over GrowSP++ and $+2.7$ over LogoSP) and OA ($84.2$) among unsupervised methods; GrowSP++ attains higher mAcc ($60.1$ vs.\ our $56.8$) but a lower mIoU ($46.6$).
Under 6-fold cross-validation, MSSP ($46.9$) improves over LogoSP ($46.3$) and is on par with GrowSP++ ($47.1$).
The S3DIS mIoU gain ($+2.6$--$2.7$) exceeds ScanNet's because S3DIS scenes are large open floor plans where objects span varying extents: multi-scale features capture this heterogeneity while spatial constraints prevent merging disconnected but feature-similar regions.

\begin{table}[H]
\centering
\caption{Results on S3DIS dataset. Left: standard Area~5 protocol (train on Areas 1, 2, 3, 4, 6; test on Area~5). Right: 6-fold cross-validation (leave-one-area-out, averaged over Areas 1--6). Best unsupervised results in \textbf{bold}. $^\dagger$Does not use pretrained 2D visual features; MSSP and the other unsupervised baselines use pretrained DINO/DINOv2 features.}
\label{tab:s3dis}
\small
\setlength{\tabcolsep}{3pt}
\begin{tabular}{l ccc ccc}
\toprule
& \multicolumn{3}{c}{Area 5} & \multicolumn{3}{c}{6-fold cross-validation} \\
\cmidrule(lr){2-4} \cmidrule(lr){5-7}
Method & OA(\%) & mAcc(\%) & mIoU(\%) & OA(\%) & mAcc(\%) & mIoU(\%) \\
\midrule
\multicolumn{7}{l}{\textit{Supervised}} \\
PointNet~\cite{Qi2017CVPR}               & 77.5 & 59.1 & 44.6 & 75.9 & 67.1 & 49.4 \\
PointNet++~\cite{qi2017pointnetplusplus}  & 77.5 & 62.6 & 50.1 & 77.1 & 74.1 & 55.1 \\
SparseConv~\cite{graham2018sparseconv}    & 88.4 & 69.2 & 60.8 & 89.4 & 78.1 & 69.2 \\
\midrule
\multicolumn{7}{l}{\textit{Unsupervised}} \\
K-means           & 21.4 & 21.2 & 8.7  & 20.0 & 21.5 & 8.8 \\
IIC~\cite{jaritz2020xmuda}               & 28.5 & 12.5 & 6.4  & 32.8 & 14.7 & 8.5 \\
PiCIE~\cite{cho2021picie}              & 61.6 & 25.8 & 17.9 & 46.4 & 28.1 & 17.8 \\
GrowSP$^\dagger$~\cite{Zhang_2023_CVPR}            & 78.4 & 57.2 & 44.5 & 76.0 & 59.4 & 44.6 \\
GrowSP++~\cite{zhang2026growsppp}          & 78.7 & \textbf{60.1} & 46.6 & 77.6 & \textbf{59.9} & \textbf{47.1} \\
PointDC~\cite{Chen_2023_ICCV}            & 54.1 & 24.1 & 22.6 & 55.7 & 37.7 & 26.0 \\
PointDC-DINOv2~\cite{Chen_2023_ICCV}     & 75.7 & 48.7 & 40.2 & 74.4 & 51.5 & 41.3 \\
LogoSP~\cite{Zhang_2025_CVPR}            & 82.8 & 55.9 & 46.5 & 79.2 & 58.0 & 46.3 \\
\textbf{MSSP (Ours)} & \textbf{84.2} & 56.8 & \textbf{49.2} & \textbf{80.5} & 59.2 & 46.9 \\
\bottomrule
\end{tabular}
\end{table}

\subsection{Qualitative Results}
\label{sec:qualitative}

Figure~\ref{fig:qualitative} shows a qualitative comparison on ScanNet (top row) and S3DIS (bottom row).
MSSP produces more coherent segments than LogoSP on both datasets, particularly on structurally complex classes and at object boundaries.
Where LogoSP often fragments large objects into inconsistent labels, MSSP maintains coherent predictions by combining spatial constraints with multi-scale context, confirming that the ``regularize-then-enrich'' interaction improves both region consistency and boundary quality.
The improvement is visually more pronounced on S3DIS than on ScanNet, consistent with the quantitative trend, as its larger scenes contain objects spanning more varying spatial extents.

\subsection{Cross-dataset Generalization}
\label{sec:generalization}

Following~\cite{Zhang_2025_CVPR}, we evaluate cross-dataset generalization by training on one dataset and testing on another without fine-tuning.
Table~\ref{tab:generalization}(a) shows ScanNet$\to$S3DIS, where MSSP outperforms LogoSP by $4.8$ mIoU, suggesting that multi-scale features from diverse scenes capture transferable structure.
Table~\ref{tab:generalization}(b) reports the reverse direction; the marginal $+0.2$ improvement is expected as S3DIS has limited diversity (271 rooms).

\begin{table}[H]
\centering
\caption{Cross-dataset generalization (mIoU\%). Best unsupervised results in \textbf{bold}.}
\label{tab:generalization}
\small
\setlength{\tabcolsep}{3pt}
\textbf{(a)} Generalization from ScanNet to S3DIS (mIoU\%)\par\smallskip
\begin{tabular}{l ccccccc}
\toprule
test on $\to$ & Area 1 & Area 2 & Area 3 & Area 4 & Area 5 & Area 6 & Mean \\
\midrule
IIC~\cite{jaritz2020xmuda}             & 3.7 & 3.8 & 3.8 & 4.0 & 3.8 & 3.7 & 3.8 \\
PiCIE~\cite{cho2021picie}             & 13.5 & 12.7 & 13.4 & 12.8 & 11.3 & 13.1 & 12.8 \\
GrowSP~\cite{Zhang_2023_CVPR}          & 24.2 & 21.9 & 26.1 & 25.0 & 23.7 & 27.9 & 24.8 \\
PointDC~\cite{Chen_2023_ICCV}          & 23.6 & 20.9 & 24.6 & 19.5 & 20.1 & 29.7 & 23.1 \\
PointDC-DINOv2~\cite{Chen_2023_ICCV}   & 33.8 & 29.6 & 33.8 & 31.7 & 32.2 & 36.9 & 33.0 \\
LogoSP~\cite{Zhang_2025_CVPR}          & 43.8 & 37.5 & 47.0 & 40.7 & 44.9 & 47.9 & 43.6 \\
\textbf{MSSP (Ours)} & \textbf{47.0} & \textbf{39.4} & \textbf{48.6} & \textbf{42.9} & \textbf{48.7} & \textbf{53.9} & \textbf{48.4} \\
\bottomrule
\end{tabular}

\vspace{6pt}
\textbf{(b)} Generalization from S3DIS to ScanNet (mIoU\%)\par\smallskip
\setlength{\tabcolsep}{2pt}
\begin{tabular}{l cccccc c}
\toprule
model trained on $\to$ & {\tiny 2/3/4/5/6} & {\tiny 1/3/4/5/6} & {\tiny 1/2/4/5/6} & {\tiny 1/2/3/5/6} & {\tiny 1/2/3/4/6} & {\tiny 1/2/3/4/5} & mean \\
\midrule
IIC~\cite{jaritz2020xmuda}             & 3.5 & 3.4 & 3.7 & 3.5 & 3.5 & 3.6 & 3.5 \\
PiCIE~\cite{cho2021picie}             & 5.6 & 5.1 & 5.0 & 5.9 & 6.0 & 5.5 & 5.5 \\
GrowSP~\cite{Zhang_2023_CVPR}          & 16.9 & \textbf{17.8} & 16.4 & 16.1 & 17.1 & 15.3 & 16.6 \\
PointDC~\cite{Chen_2023_ICCV}          & 10.3 & 10.1 & 10.4 & 8.4 & 10.0 & 9.9 & 9.9 \\
PointDC-DINOv2~\cite{Chen_2023_ICCV}   & 16.7 & 15.4 & 16.4 & \textbf{17.6} & 14.3 & 14.6 & 15.8 \\
LogoSP~\cite{Zhang_2025_CVPR}          & 16.9 & 16.9 & 16.8 & 16.8 & 16.7 & \textbf{17.0} & 16.9 \\
\textbf{MSSP (Ours)} & \textbf{17.4} & 17.1 & \textbf{17.1} & 16.8 & \textbf{18.0} & 16.2 & \textbf{17.1} \\
\bottomrule
\end{tabular}
\end{table}

\subsection{Ablation Study}
\label{sec:ablation}

We conduct ablation experiments on S3DIS (Area~5 test), incrementally adding each component to the LogoSP~\cite{Zhang_2025_CVPR} baseline.
Table~\ref{tab:ablation} reports the results.

\begin{table}[H]
\centering
\caption{Ablation study on S3DIS Area~5. SC: Spatially-constrained clustering. MS: Multi-scale spectral analysis. Best in \textbf{bold}.}
\label{tab:ablation}
\small
\begin{tabular}{l cc ccc}
\toprule
Setting & SC & MS & OA(\%) & mAcc(\%) & mIoU(\%) \\
\midrule
LogoSP baseline &     &     & 82.8 & 55.9 & 46.5 \\
\midrule
+SC only           & \checkmark &     & 83.2 & 51.7 & 45.3 \\
+MS only           &     & \checkmark & 80.8 & \textbf{57.4} & 45.3 \\
+SC + MS (Ours) & \checkmark & \checkmark & \textbf{84.2} & 56.8 & \textbf{49.2} \\
\bottomrule
\end{tabular}
\end{table}

\begin{figure}[t]
\centering
\includegraphics[width=\textwidth]{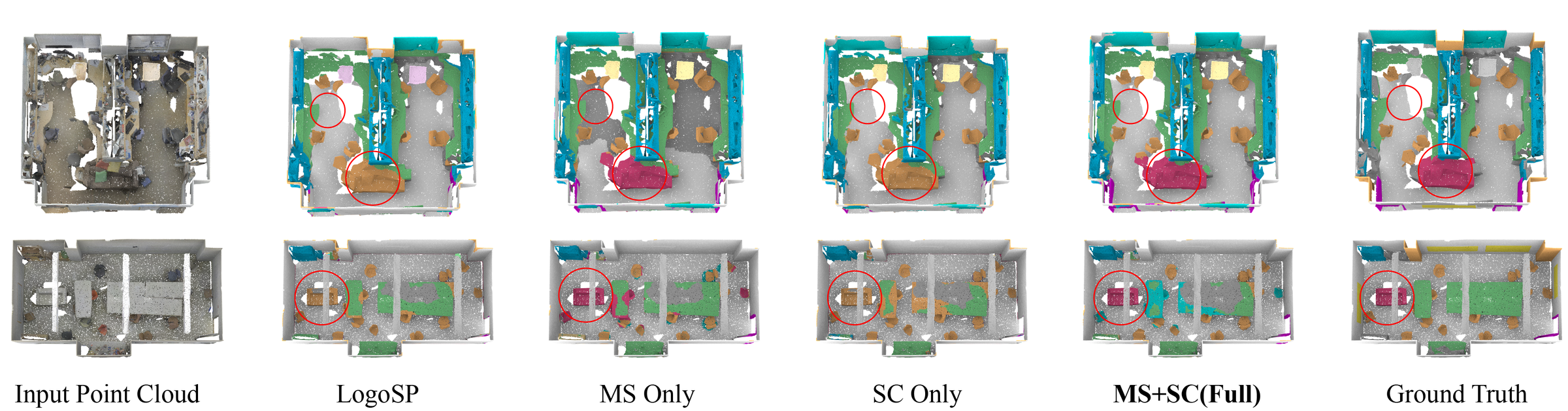}
\caption{Ablation visualization on S3DIS. Neither MS nor SC alone improves over the baseline, but their combination produces markedly better segmentation.}
\label{fig:ablation}
\end{figure}

Table~\ref{tab:ablation} reveals a pronounced synergy between the two components.
Applied in isolation, neither improves over the baseline: SC alone reaches $45.3$ mIoU and MS alone $45.3$, both below the $46.5$ of the LogoSP baseline.
Their combination, however, attains $49.2$ mIoU---an improvement of $+2.7$---demonstrating that the components are complementary rather than redundantly beneficial.
The single-component runs also expose a trade-off between the two metrics: SC alone lifts OA to $83.2$ at the cost of mAcc ($51.7$), whereas MS alone raises mAcc to $57.4$ but lowers OA to $80.8$.
Only their joint use recovers both axes while gaining mIoU, consistent with a \emph{regularize-then-enrich} interaction that we analyze next and visualize in Figure~\ref{fig:ablation}.

Table~\ref{tab:ablation} reveals a ``regularize-then-enrich'' interaction.
\textbf{Why MS alone falls short ($-1.2$ mIoU):} The $3\times$ expanded descriptors are richer, yet without a spatial-coherence prior the additional dimensions do not by themselves produce more coherent pseudo-labels or better final segmentation.
\textbf{Why SC alone falls short ($-1.2$ mIoU):} With single-scale features, the spatial adjacency constraint lacks the discriminative signal needed to form cleaner primitives, so the structural prior on its own does not improve over the baseline.
\textbf{Why their combination works ($+2.7$ mIoU):} MS supplies the discriminability to separate adjacent-but-distinct regions (e.g., a \emph{chair} vs.\ a \emph{table}), while SC enforces spatial coherence so that the richer descriptors converge to consistent connected regions. We verify this directly on the pseudo-labels: spatial coherence---the fraction of 6-connected adjacent voxel pairs that share a label---rises from $90.9\%$ for the LogoSP baseline to $92.5\%$ for the full model, and the number of effective primitives (clusters covering at least $1\%$ of points) is largely retained ($13 \to 12$), indicating that SC supplies the spatial-coherence prior while MS preserves the underlying semantic diversity.
Consistent with this, the full model improves over the LogoSP baseline on all three axes simultaneously (OA $82.8{\to}84.2$, mAcc $55.9{\to}56.8$, mIoU $46.5{\to}49.2$), confirming that the synergy does not trade one metric for another.

\subsection{Component Sensitivity Analysis}
\label{sec:sensitivity}

\textbf{Clustering granularity $\mathcal{K}$.}
Table~\ref{tab:scale_ablation} varies the clustering scales on ScanNet.
Among single scales, the medium granularity $\{40\}$ performs best ($34.1$ mIoU), as it roughly aligns with whole-object extent, while the finer $\{80\}$ ($33.2$) and the coarser $\{20\}$ ($31.4$) lose discrimination at either end of the spectrum.
Pairwise combinations improve only marginally over the best single scale ($34.4$--$34.8$ mIoU), yet adding the third scale yields a clear jump to $36.2$ mIoU and $51.3$ mAcc---a $+1.4$ mIoU gain over the best pair.
This non-linear trend indicates that the three granularities are largely complementary: fine clusters capture object parts, medium clusters capture whole objects, and coarse clusters capture scene-level regions, and only their union covers the full range of spatial extents present in indoor scenes, which the spatial-coherence prior then consolidates into coherent primitives.

\begin{table}[H]
\centering
\caption{Multi-scale granularity ablation on ScanNet. All variants use the same experimental setup, varying only the set of KMeans scales $\mathcal{K}$. Best in \textbf{bold}.}
\label{tab:scale_ablation}
\small
\begin{tabular}{l c c c}
\toprule
$\mathcal{K}$ & Dim & mIoU(\%) & mAcc(\%) \\
\midrule
$\{80\}$   & 384  & 33.2 & 46.9 \\
$\{40\}$   & 384  & 34.1 & 49.9 \\
$\{20\}$   & 384  & 31.4 & 46.3 \\
$\{80,40\}$ & 768  & 34.4 & 50.5 \\
$\{40,20\}$ & 768  & 34.7 & 48.6 \\
$\{80,20\}$ & 768  & 34.8 & 49.8 \\
$\{80,40,20\}$ & 1152 & \textbf{36.2} & \textbf{51.3} \\
\bottomrule
\end{tabular}
\end{table}

\section{Conclusion}
\label{sec:conclusion}

We presented MSSP for unsupervised 3D point cloud semantic segmentation.
Our core contribution is multi-scale spectral analysis that builds enriched superpoint descriptors at multiple clustering granularities, with spatially-constrained clustering as structural regularization.
A key finding is that multi-scale features alone do not improve final segmentation, yet become highly effective combined with spatial constraints, yielding the best mIoU on the S3DIS Area-5 and ScanNet benchmarks among unsupervised methods, albeit at a higher computational cost than the single-scale baseline.
Future work includes exploring adaptive scale selection and extending the spatial regularization to the backbone training phase.

\subsubsection*{Disclosure of Interests.}
The authors have no competing interests to declare that are relevant to the content of this article.

\bibliographystyle{splncs04}
\bibliography{references}

\end{document}